\documentclass[fleqn,10pt]{wlscirep}
\usepackage[utf8]{inputenc}
\usepackage[T1]{fontenc}
\usepackage{array}
\usepackage{amsmath}
\usepackage{float}

\title{Machine Learning based CVD Virtual Metrology in Mass Produced Semiconductor Process}

\author[1,*]{Yunsong Xie}
\author[1]{Ryan Stearrett}
\affil[1]{Samsung Austin Semiconductor Company, Austin, TX, 78754, USA}

\affil[*]{yunsong.xie@samsung.com}

\begin{abstract}

A cross-benchmark has been done on three critical aspects, data imputing, feature selection and regression algorithms, for machine learning based chemical vapor deposition (CVD) virtual metrology (VM). The result reveals that linear feature selection regression algorithm would extensively under-fit the VM data. Data imputing is also necessary to achieve a higher prediction accuracy as the data availability is only $\sim$70\% when optimal accuracy is obtained. This work suggests a nonlinear feature selection and regression algorithm combined with nearest data imputing algorithm would provide a prediction accuracy as high as 0.7. This would lead to 70\% reduced CVD processing variation, which is believed to will lead to reduced frequency of physical metrology as well as more reliable mass-produced wafer with improved quality.

\end{abstract}

\begin{document}
	
	\flushbottom
	\maketitle
	\thispagestyle{empty}

	\section*{I. Introduction}
	
	Semiconductors have been an essential component of the modern society for decades. It provides critical products such as computing, communicating, and data storage devices that plays major roles in healthcare \cite{fratini2020charge, de2017solution}, entertainment \cite{deen201720, zhu2019organic}, military systems \cite{khandelwal2017performance, kanarik2018ale, campbell2018high}, transportation \cite{schulz2016power, csizmadia2019power, kumar2018performance}, clean energy \cite{miyata2020hybrid, johns2019room, zhao2016nanosecond}, and countless other applications \cite{kim2017materials, yu2016metal, rim2017interface, xu2017highly, zhang2016alignment}. Chemical Vapor Deposition (CVD) is one of the critical and commonly used processes in semiconductor fabrication processes. It provides the capability of conformally depositing a layer of insulator on top of the wafer. \cite{jones2008cvd, gleason2015cvd} In mass produced semiconductor fabrication process, to ensure that CVD process is highly accurate, stable and reliable, one has to implement the Advanced Process Control (APC) system to the wafer processing. A negative feedback loop is integrated in the APC system to minimize the process variability caused by the natural drifting of the instrument parameters. The basic concept of APC is to adjust the process condition according to the measurement of processing result. \cite{moyne2000run} Run-to-run (R2R) is one of the most sophisticated and well performed process control algorithm for producing the accurate processing result. R2R not only uses the negative feedback loop like normal APC system, but also builds a process model to compensate the processing parameter drifts, incoming product variation, and other disturbances. \cite{wan2017gaussian} However, in order to achieve a better controlled process, adequate metrology systems need to be employed to monitor every wafer and update the process model. The increased metrology activities as well as extra investment for metrology tools would not only prolong the wafer processing time but also dramatically increase the cost of the process.

	As a solution to the cost/quality dilemma, virtual metrology (VM) has been developed. VM uses the data collected in the tool sensors, called Fault Detection and Classification (FDC), to predict the metrology result that can be used in R2R process control system. \cite{wan2017gaussian, purwins2013regression, jebri2017virtual} VM is able to dramatically reduce the necessary physical metrology frequency thus reducing the cost and wafer processing time. Furthermore, VM is also able to improve product quality by reducing processing variation. As a metric, the performance of a selected VM is evaluated by the prediction accuracy as it directly impacts the effectiveness of R2R process control system. \cite{wan2017gaussian, purwins2013regression, jebri2017virtual}

	Many works have been dedicated to improve the accuracy of VM by implementing different predicting regression algorithms. Among them, the predicting regression algorithms can be roughly divided into three categories:
	\begin{itemize}
		\item Linear regression. It is one of the simplest and most widely used algorithm for predicting the VM result. Although this algorithm does not provide the best fitting accuracy, it does provide an interpretable model that can be beneficial to the engineer for gaining an easy-to-understand processing knowledge. The most often studied linear regression algorithms include Multiple Linear Regression \cite{purwins2013regression, kwon2008data} (MLR), Least Absolute Shrinkage and Selection Operator (LASSO) regression \cite{yang2019structure, pampuri2011multilevel, pampuri2012multistep, arima2019applications}, Partial Linear Square Regression (PLS) \cite{purwins2013regression, purwins2013regression, susto2011virtual, wu2016virtual, besnard2012virtual, ferreira2011virtual, arima2019applications, roh2018development}, Support Vector Regression (SVR) \cite{arima2019applications, lenz2013data, purwins2013regression, lenz2013virtual} and Ridge Linear (RL) regression \cite{purwins2013regression}, etc.

		\item Decision trees based regression. Decision tree regression provides a good balance between interpretability and fitting accuracy comparing to linear and network based regressions. It includes standard decision trees regression \cite{lenz2013virtual, yang2019structure}, Random Forest regression \cite{yang2019structure}, Gradient Boosting (GB) regression \cite{roeder2014feasibility} and Tree Ensemble regression \cite{ferreira2011virtual, quinlan2014c4, besnard2012virtual}, etc.
		
		\item Network based regression. Network based algorithm has little interpretability, but it is able to provide very high fitting accuracy and computation efficiency. Neural Network (NN) regression is one of the most popular algorithm in this category. \cite{jia2018adaptive, susto2011virtual, susto2013virtual, kwon2008data} Besides NN, Group Method of Data Handling (GMDH) network \cite{jia2018adaptive} and Gaussian Bayesian network regression \cite{yang2019structure} have also been considered in the past VM works.
	\end{itemize}
	
	Besides selecting a better regression algorithm, feature selection algorithm is another important field for general machine learning researches. \cite{miao2016survey, masoudi2019featureselect} Some VM studies use engineering experience to determine which features to be included in the prediction. \cite{jia2018adaptive, kurz2014sampling, wan2017gaussian} Plenty of other studies use variety of machine learning algorithm to determine the features to be considered in the prediction, such as Principle Component Analysis \cite{susto2011virtual, susto2013virtual, arima2019applications}, PLS \cite{roh2018development}, Random Forest \cite{yang2019structure}, Statistic \cite{jia2018adaptive}, Support vector machine \cite{kwon2008data}.
	
	The third aspect of achieving a accurate prediction result is data imputing, which is the method of handling the missing data. Despite missing data being very common in the raw manufacturing FDC dataset, and imputing data is a very important subject for the machine learning society \cite{mccoy2018variational, beaulieu2017missing, efron1994missing}, there is little discussion in the previous VM studies about how imputing the missing data would impact the final prediction accuracy. Many studies simply discard the samples that have missing values. \cite{yang2019structure, susto2011virtual, jebri2017virtual, besnard2012virtual, ferreira2011virtual, susto2013virtual}
	
	This work is the first attempt to cross-benchmark the CVD VM performance by evaluate all three aspects: algorithm, feature selection, and data imputing. In algorithm aspect, six different methods from all three main categories are included in the studies. Linear Least Squares (LLS), PLS, linear SVR and Bayesian linear (BL) regressions are included for linear category. GB regression for decision tree based category, and NN regression are included for network based category. In feature selection aspect, since all linear algorithms and GB generate feature importance information, these algorithms would gradually reduce the considered features by selecting the most important ones. However, NN algorithm, however, will use the feature selection function of GB since both them are nonlinear algorithms. In data imputing aspect, some of the most popular methods including random, K-nearest neighbor (KNN), AutoRegressive Integrated Moving Average (ARIMA), nearest, and Random Forest are considered. The result reveals that nonlinear feature selection and regression algorithm combined with nearest data imputing algorithm would provide a testing accuracy as high as 0.7, which means that 70\% of the processing variation is predicted by the proposed method.  It is believed by appropriately adjust the processing according to the prediction a reduced frequency of physical metrology as well as 70\% lower processing variable \cite{zeng2012statistical, tanaka2013prediction, tsutsui2019virtual}, that would eventually result in lower production cost and much better quality control for semiconductor mass production processing.

	\section*{II. Methodology}
	
	\subsection*{II.1. Data overview}
	
	To avoid revealing proprietary information, no details about the features will be disclosed in this manuscript. All features are normalized to be within the range between 0 and 1. The entire data considered in this study are divided into 3 datasets, which are training, development and testing sets. The sample size ratio among these three datasets are 70\%, 15\%, and 15\%. Training and development datasets are used to train the prediction regression algorithms and hyperparameters selection, respectively. Testing dataset is used to do the final accuracy benchmark.

	\subsection*{II.2. Missing value imputation}

	\begin{figure}[ht]
		\centering
		\includegraphics[width=70mm]{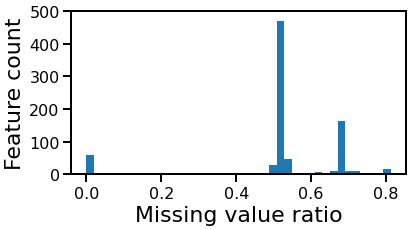}
		\caption{Histogram plot of missing data ratio for features of CVD process in mass production. X axis is the binned missing data ratio. Y axis is the feature count that belongs to given bin.}
		\label{fig:Missing_stats}
	\end{figure}
	
	The CVD process under investigation has 715 features that include both continuous numerical data such as temperature, gas flow rate, electrical voltage, RF frequency, and gas pressure, as well as categorical data such as manufacturing/metrology tool IDs and chambers. Unlike the ideal scenario in laboratory fabrication, the mass production tool dataset always contains a large portion of missing data. \cite{yang2019structure, susto2011virtual, besnard2012virtual, ferreira2011virtual} As shown in Figure \ref{fig:Missing_stats}, there are 500 features in the entire dataset that have around 50\% data availability. Only about 60 features that contain >95\% data availability. The missing data is a combination result of complexity of the process, cost of collecting, processing, and storing high volume data entries. This missing data may not be one of the biggest concerns according to human's empirical decision, but the through imputing, it could offer valuable information in the machine learning process. \cite{bertsimas2017predictive, sovilj2016extreme} Unlike the majority of prior virtual metrology studies that simply select the datasets that have high availability and eliminate the samples that do not have the all selected features data, all 715 features are included into the machine learning algorithm consideration. Five common data imputing algorithms are utilized in the cross-benchmark: 
	
	\begin{enumerate}
		\item ARIMA: Time series imputing algorithm that combines autoregressive, degree of differencing, and moving-averaging models. The imputing is applied to each features and production tool separately.
		\item KNN: Imputing algorithm that fills the missing data with nearest neighbor averaging. During imputing, all features from a selected tool is fed into the algorithm at one time.
		\item Nearest: The missing data entry takes the last available data entry that collected for the same tool and sensor.
		\item Random: Randomly assign the missing data entry with a value ranging from 0 to 1.
		\item Random Forest: Uses Random Forest algorithm to impute the missing data. Similar to KNN, all features from a selected tool is fed into the algorithm at one time.
	\end{enumerate}

	\subsection*{II.3. Feature selection and regression}
	It has been well recognized in the previous virtual metrology studies that feature selection has a huge impact on the prediction accuracy. When too many features are included, the model will be over-fitted. The concept of feature selection is to use a regression algorithm that can generate the importance metric to each input feature. The selected features are the ones have highest importance metrics. 5 out of 6 feature selection algorithms (LN, PLS, BR, SV, and GB) considered in this manuscript uses the same feature selection and regression algorithms. However, NN regression algorithm uses the GB algorithm for feature selection because NN does not produce feature importance information. Table 1 shows the relationship between the feature selection and regression algorithms that are used in the manuscript.
	
	\begin{table}[h]
		\centering
		\caption{Lookup table for feature selection and regression algorithms.}
		\label{table:feature_selection_table}
		\begin{tabular}{|c|c|c|}
			\hline
			\textbf{Linearity type}	& \textbf{Feature selection algorithm}	& \textbf{Regression algorithm}		\\ [0.25ex] \hline \hline 
			Linear		& Linear least squares		& Linear least squares		\\ \hline
			Linear		& Partial least squares		& Partial least squares 	\\ \hline
			Linear		& Bayesian Ridge 			& Bayesian Ridge 			\\ \hline
			Linear		& Support vector 			& Support vector 			\\ \hline
			Non-linear	& Gradient Boosting 		& Gradient Boosting 		\\ \hline
			Non-linear	& Fully connected neural network  			& Gradient Boosting 		\\ \hline

		\end{tabular}
	\end{table}
	
	The other important input of the feature selection algorithm is the number of the selected/output features. In this study, the feature selection algorithm is to be applied to all the features multiple times. Each time the number of selected feature is reduced by a ratio between 0.08 and 0.11 depending on the number of input features (NIF). A less aggressive feature reduction ratio is chosen at a lower NIF in order to gain a finer NIF to prediction accuracy dependence when remained features are the ones that are more relevant to the prediction result.

	\section*{III. Results}
	
	The cross-benchmark is to have all six regression algorithms be applied on the three different datasets. Each dataset is imputed by 5 different imputing algorithms. The calculations have been done with variety of NIF. For given input features and regression algorithm, the following calculation procedure was executed:
	
	\begin{itemize}
	\item Use the given regression algorithm to fit the training dataset.
	\item Optimize the hyperparameters of regression algorithm based on the predicted accuracy result on the development dataset.
	\item Achieve the final regression algorithm.
	\item Use the feature selection algorithm to output the selected features to be used in the next iteration.
	\end{itemize}
	
	\begin{figure}[ht]
		\centering
		\includegraphics[width=175mm]{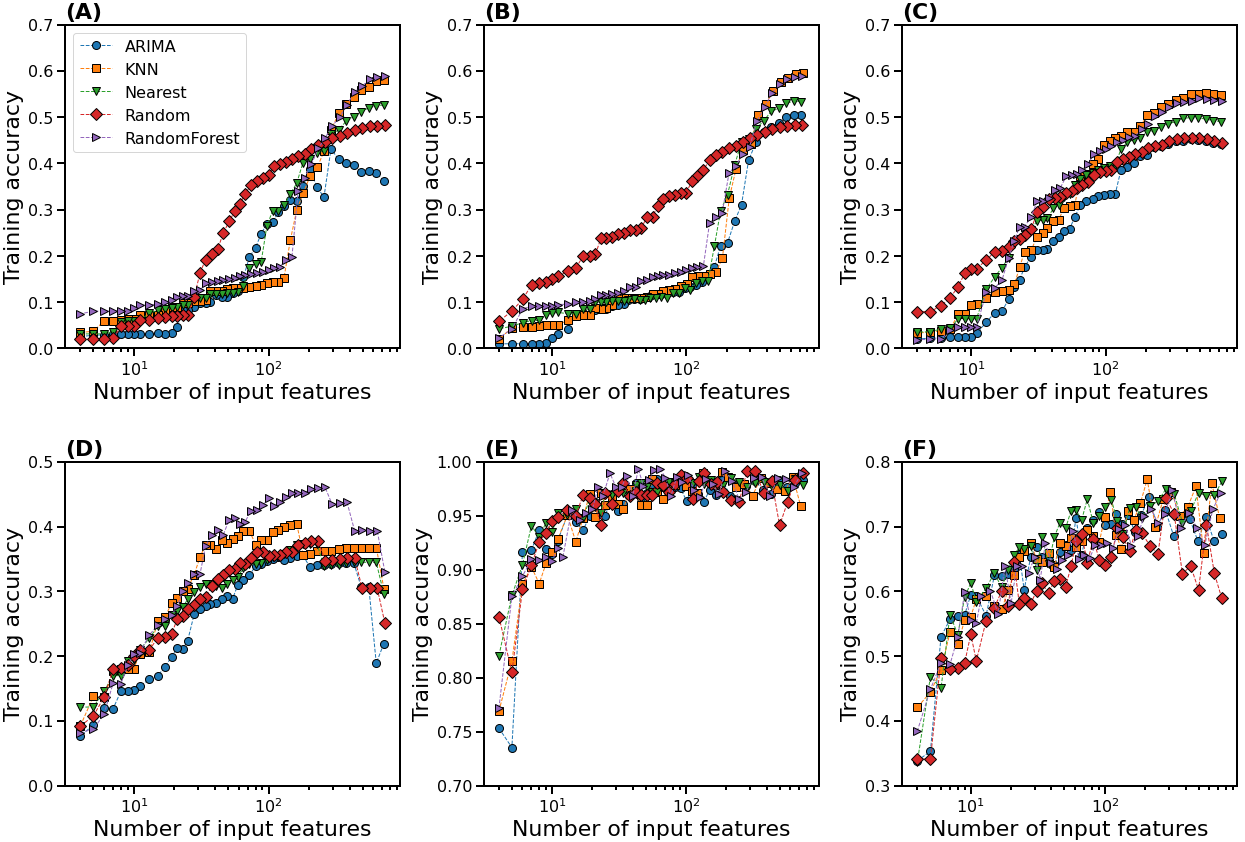}
		\caption{The accuracy of the regression on training dataset with respect to input number of features. Each figure represents a different regression algorithm: (A) LN, (B) PLS, (C) BR, (D) SV, (E) GB, and (F) NN. All 5 imputing methods are shown in each figure. X axis, plotted in logarithmic scale, is the input number of features.}
		\label{fig:Accuracy_summary_training}
	\end{figure}

	The training accuracy dependence on the NIF of the finalized algorithm on training dataset is shown in Figure \ref{fig:Accuracy_summary_training}. Each data point in the figure represents a completion of a series of calculation procedure. The accuracy data is R$^2$ of the linear fitting between the measured and predicted metrology data. All non-linear algorithms, GB and NN (Figure \ref{fig:Accuracy_summary_training} (E,F)), produce better fitting performance comparing to the linear algorithms (Figure \ref{fig:Accuracy_summary_training} (A-D)). The reason is that the linear algorithms constantly under-fit the data since they do not have enough degrees of freedom to fully characterize the variation of metrology data. This shows that in order to produce a meaningful metrology prediction, a non-linear regression algorithm is mandatory despite linear algorithm was widely used in some of the prior virtual metrology investigation.
	
	Among the 6 regression methods, GB produces the training accuracy of beyond 0.95, shown in Figure \ref{fig:Accuracy_summary_training}(E), when NIF is more than 50. The accuracy decreases dramatically as the number of input reduces. This shows that the information, which is needed to accurately predict virtual metrology results, starts to lose when number of features less than $\sim$ 50. The rate of losing information accelerates when NIF further reduces to less than  $\sim$ 50, which means that the later-eliminated features enclose more valuable information for an accurate prediction. NN algorithm produces training accuracy from 0.6 to 0.8, according to Figure \ref{fig:Accuracy_summary_training}(F). Despite lower absolute value, the training accuracy dependence on NIF shows a similar overall trend, in which the training accuracy starts to decrease with a faster rate as the NIF reducing to less than $\sim$ 50.

	\begin{figure}[ht]
		\centering
		\includegraphics[width=175mm]{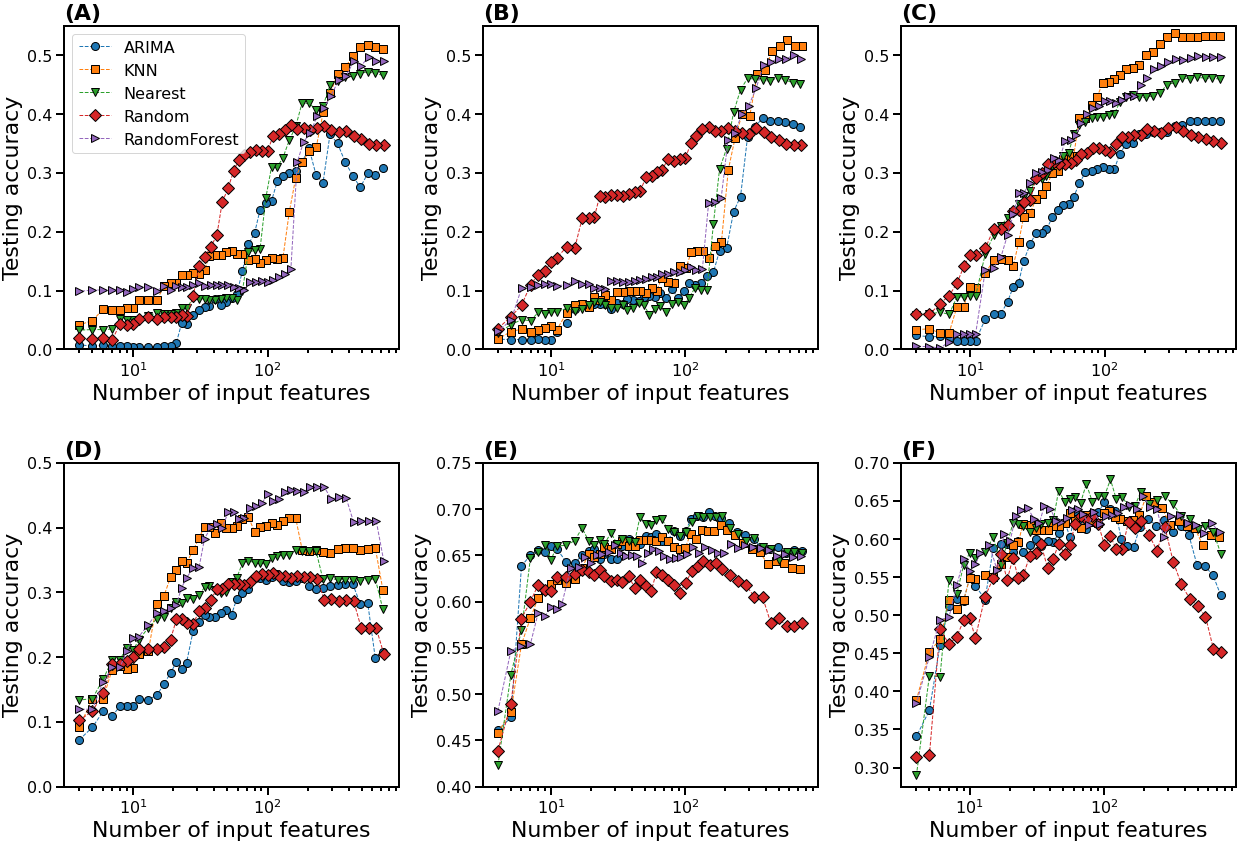}
		\caption{The accuracy of the regression on testing dataset with respect to input number of features. Each figure represent a different regression algorithm: (A) LN, (B) PLS, (C) BR, (D) SV, (E) GB, and (F) NN. All 5 imputing methods are shown in each figure. X axis, plotted in logarithmic scale, is the input number of features.}
		\label{fig:Accuracy_summary_testing}
	\end{figure}

	Using Figure \ref{fig:Accuracy_summary_training}, one could argue that NN algorithm has a worse performance prediction comparing to GB. However,the accuracy dependence on the NIF achieved with testing dataset, shown in Figure \ref{fig:Accuracy_summary_testing}, shows a very different result. Even though GB algorithm, shown in Figure \ref{fig:Accuracy_summary_training}(E), provides a highly accurate prediction in the training dataset, it only makes nearly 0.7 in the best scenario with testing dataset. This means that the prediction of GB algorithm for training dataset is a result of over-fitting. The accuracy on the testing dataset provided by NN algorithm, shown in Figure \ref{fig:Accuracy_summary_testing}(F), is also close to 0.7 in the best scenario, which is close to the accuracy of training dataset with the same algorithm. It indicates that the fitting capability of NN algorithm on the training dataset is neither over-fit nor under-fit. Both nonlinear algorithms, GB and NN, show strong dependence on NIF. The highest accuracy for NN peaks at NIF around 150 regardless what imputing method is used. Such observation confirms that feature selection is necessary in the machine learning based virtual metrology study. The linear algorithms all show inferiority in testing accuracy comparing to nonlinear algorithms because of under-fitting, which is consistent with the conclusion that has been drawn above. 
	
	The significance of prediction on testing dataset is that since this dataset has never been seen by the algorithm before, the prediction accuracy represents the ratio of random variation that can be explained by the trained algorithm. If the CVD process can be control according to the algorithm prediction, the processing variation can be reduced by the amount of the accuracy. \cite{zeng2012statistical, tanaka2013prediction, tsutsui2019virtual}. In the best scenario, a nearly 70\% of CVD processing variation can be expected as Figure \ref{fig:Accuracy_summary_testing}(E) and (F) suggest.
	
	\begin{figure}[ht]
		\centering
		\includegraphics[width=180mm]{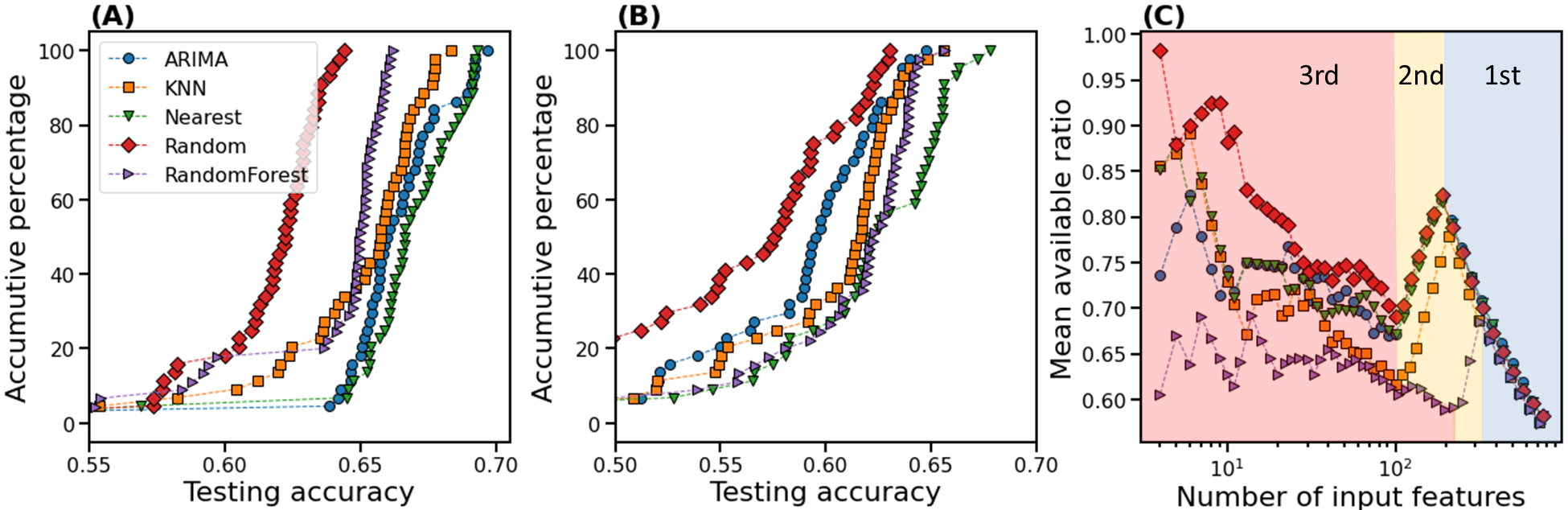}
		\caption{The cumulative plot for testing accuracy with all 5 imputing methods for GB (A) and NN (B) algorithms. Mean  data available ratio (MDAR) dependence on number of input features (NIF) for both  GB and NN algorithms (C).}
		\label{fig:impute_feature_analysis}
	\end{figure}
	
	The imputing method also proves itself to be an critical factor for testing accuracy. The clear worst imputing method is random imputing, which is reasonable because it offers nothing but noise to the dataset. The remaining 4 imputing methods actually produce similar accuracy while using the nonlinear algorithm, GB and NN. Figure \ref{fig:impute_feature_analysis} offers more detailed analysis on the input features for GB and NN algorithms to highlight the difference among the 5 imputing algorithms. Accumulative plots for testing accuracy of both GB and NN algorithms are shown in Figure \ref{fig:impute_feature_analysis}(A) and \ref{fig:impute_feature_analysis}(B), respectively, where all 5 imputing methods are included. While Random imputing methods provides the worse accuracy for both regression algorithm, Random Forest, ARIMA, and KNN imputing give intermediate accuracy. The data imputed by Nearest imputing algorithm  leads to the best testing accuracy with both regression algorithms. A deeper analysis on the level of imputing has been done by investigating the data missing level of the input features for both GB and NN regression algorithms. Figure \ref{fig:impute_feature_analysis}(C) illustrates mean data available ratio (MDAR) for all input features for both GB and NN regression algorithms (both of these two regression algorithms use GB as the feature selection algorithms so that the input features are the same). MDAR is calculated by:
	
	\begin{equation} \label{data_available_eqp}
	MDAR = \frac{\displaystyle\sum_{i=1}^{M} \displaystyle\sum_{j=1}^{N} A_{ij}}{MN}
	\end{equation} 
		
	where M and N are the number of input features and samples respectively. $A_{ij}$ is an indicator of whether the data at $i$-th feature and $j$-th sample available. If the data is available, $A_{ij}$ is 1, otherwise, the data has to be imputed and $A_{ij}$ becomes 0. The dependence of MDAR on NIF for all 5 imputing methods are similar and can be roughly divided into three stages. The 1st stage is when the number of features is around 200 (300 for Random Forest algorithm), where MDAR increase as NIF reduces. It indicates that some of the features with very low availability is being eliminated in this stage. The 2nd stage appears when the NIF is between 100 and 200 (200 and 300 for Random Forest algorithm). MDAR reduces drastically at this stage as NIF reduces, which is caused by some of the features with high availability but does not impact the metrology result is being eliminated in this stage. The 3rd stage occurs when NIF less than 100 (200 for Random Forest). In this stage, MDAR keeps a rising trend as the features being eliminated. Such trend only ends when NIF reduces to less then 10, where the test accuracy hits a cliff, meaning that the remaining feature has lost most of the information that would impact metrology result. The best testing accuracy is also achieved around 100 features (200 for Random Forest algorithm) that aligns well with the boundary of stage 2nd and 3rd. This means that when input feature is around 100, the imputed information resulted from features with low availability plays critical role to boost the test accuracy.

	\section*{IV. Conclusions}
	
	This study is the first trial in the virtual metrology community to include multiple popular data imputing, feature selection, and regression algorithms in the cross-benchmark. The result shows that non-linear algorithm should be used for both feature selection and regression algorithms since linear algorithm would under-fit the data extensively. Both non-linear regressions under investigation, GB and NN, produce similar testing accuracy. This study also prove that random imputing and nearest imputing give the worse and best testing accuracy results, respectively. Imputed data play an important role for the prediction as mean data available ratio is at a local minimal point when optimal testing accuracy appears. In summary, this work paves the foundation for future virtual metrology development. Furthermore, it recommends that a combination of nearest imputing method and non-linear feature selection and regression algorithm should be used in combination to achieve a optimal accuracy. Such combination would product a prediction accuracy on testing dataset of 0.7, which means 70\% of the random processing variation can by reduced on the current mass-produced wafer.

	\section*{V. Acknowledgements}
	
	The author would thank the support from Samsung Austin Semiconductor company.

	\section*{Author contributions statement}

	Y.X. conceived the main idea, conducted the data requisition, analyzed the result and reviewed the manuscript. R. S. analyzed the result and reviewed the manuscript.

	\bibliography{sample}

\end{document}